%% file: main.tex
\pdfoutput=1

\documentclass[11pt]{article}

\usepackage[]{acl}

\usepackage{times}
\usepackage{latexsym}

\usepackage[T1]{fontenc}

\usepackage[utf8]{inputenc}

\usepackage{microtype}

\usepackage{comment}              
\usepackage{xparse}
\usepackage{color}
\usepackage{graphicx}
\usepackage{multirow}
\usepackage{makecell}
\usepackage{booktabs} 
\usepackage{ulem} 

\NewDocumentCommand{\heng}{mO{}}{\textcolor{red}{\textsuperscript{\textit{Heng}}\textsf{\textbf{\small[#1]}}}}
\NewDocumentCommand{\vicki}{mO{}}{\textcolor{blue}{\textsuperscript{\textit{Vicki}}\textsf{\textbf{\small[#1]}}}}
\NewDocumentCommand{\qiusi}{mO{}}{\textcolor{blue}{\textsuperscript{\textit{Qiusi}}\textsf{\textbf{\small[#1]}}}}

\title{EA$^2$E: Improving Consistency with Event Awareness\\ for Document-Level Argument Extraction}

\author{Qi Zeng\thanks{~~Equal contribution.},
~Qiusi Zhan$^{*}$,
Heng Ji \\
University of Illinois Urbana-Champaign   \\
\texttt{\{qizeng2, qiusiz2,hengji\}@illinois.edu } \\
}

\begin{document}
\maketitle

\input{000abstract}
\input{100introduction}

\input{300method}
\input{400experiment}

\input{500related}

\input{600conclusion}

\section*{Acknowledgement}
We thank the anonymous reviewers helpful suggestions. 
This research is based upon work supported by U.S. DARPA AIDA Program No. FA8750-18-2-0014, U.S. DARPA KAIROS Program No. FA8750-19-2-1004, and DARPA INCAS Program No. HR001121C0165. The views and conclusions contained herein are those of the authors and should not be interpreted as necessarily representing the official policies, either expressed or implied, of DARPA, or the U.S. Government. The U.S. Government is authorized to reproduce and distribute reprints for governmental purposes notwithstanding any copyright annotation therein.

\bibliography{custom}
\bibliographystyle{acl_natbib}

\end{document}

%% file: 000abstract.tex
\begin{abstract}
Events are inter-related in documents.  Motivated by the one-sense-per-discourse theory, we hypothesize that a participant tends to play consistent roles across multiple events in the same document. However recent work on document-level event argument extraction models each individual event in isolation and therefore causes inconsistency among extracted arguments across events, which will further cause discrepancy for downstream applications such as event knowledge base population, question answering, and hypothesis generation.  In this work, we formulate event argument consistency as the constraints from event-event relations under the document-level setting.  To improve consistency we introduce the Event-Aware Argument Extraction (EA$^2$E) model with augmented context for training and inference. Experiment results on WIKIEVENTS and ACE2005 datasets demonstrate the effectiveness of EA$^2$E compared to baseline methods\footnote{Our code is released at \url{https://github.com/ZQS1943/DOCIE}}.

\end{abstract}

%% file: 100introduction.tex
\section{Introduction}

\input{figures/example}

Document-level Event Argument Extraction aims at identifying arguments and their roles for multiple events in a document.  
It is a practically more useful but more challenging task than sentence-level Argument Extraction~\cite{DBLP:conf/naacl/NguyenCG16, DBLP:conf/emnlp/WaddenWLH19, DBLP:conf/acl/LinJHW20} because in a typical long input document events usually scatter across multiple sentences and are inherently connected.

Multiple events in one document are usually interconnected, and thus the arguments are under certain consistency constraints.
In Figure~\ref{fig:example}, the roles of the shared argument \textit{Ahmad Khan Rahami} in multiple events are constrained because the \textit{Attacker} in the \textit{DetonateExplode} event is likely to be the \textit{IdentifiedRole} in \textit{IdentifyCategorize} event, the \textit{Detainee} in the  \textit{ArrestJailDetain} event, as well as the \textit{Defendant} in \textit{ChargeIndict} event.
Motivated by the one-sense-per-discourse theory~\cite{DBLP:conf/naacl/GaleCY92} that mentions of an ambiguous word usually tend to share the same sense in a given discourse,
we hypothesize that \textbf{a participant tends to play consistent roles across multiple events in the same document}.
However, previous work such as~\cite{DBLP:conf/naacl/LiJH21, DBLP:conf/eacl/DuRC21, DBLP:conf/acl/YangS000W20} on document-level event argument extraction focuses on modeling each event independently and ignores the relation between events, and thus the extracted arguments of multiple events may violate the constraints from event-event relations. We call this inconsistency phenomenon. 

Though received much attention in various areas like Abstractive Summarization~\cite{DBLP:conf/eacl/NanNWSZZMX21,DBLP:conf/naacl/ZhuHXZZHJ21, DBLP:conf/emnlp/KryscinskiMXS20} and Question Answering~\cite{DBLP:journals/corr/abs-2104-08202, DBLP:conf/acl/RibeiroGS19}, the inconsistency phenomenon addressed in previous research focuses on factual consistency instead of self-contained consistency as in document-level argument extraction.
We approach this problem with inspiration from human behavior: while reading, humans subconsciously infer the event-event relations and correctly identify the event arguments under the perceived constraints. 
Therefore, we refer consistent argument extraction to applying the underlying Event-Event Relations as constraints in multi-event argument extraction.

An intuitive solution to improve consistency is to incorporate explicit Event-Event Relations into the extraction process as additional input.
However, the underlying event-event relations are hard to identify and classify due to the lack of reliable  
resources as supervision signals, especially when the arguments are unknown.
In addition, precise event-event relations may not be necessary for argument extraction when the implicit connections can already well support argument extraction.

To avoid explicit modeling of event-event relations, 
we label the arguments of other events in the context as an implication of event-event relations.
We propose an \textbf{Event-Aware Argument Extraction (EA$^2$E)} model, which incorporates alignment-enhanced training and iterative inference.
When extracting arguments, the context can be self-augmented by tagging the argument labels of other events.
Alignment-enhanced training implicitly introduces event awareness by pulling close the argument representation distributions under regular context and augmented context, where ground-truth argument labels of neighboring events are labeled. 
Iterative inference explicitly encourages event awareness by augmenting the context with the extracted arguments from the last inference iteration.
The advantage of this method is that no predefined Event-Event Relation is required, nor event schema.
Our experiments on WIKIEVENTS and ACE2005 datasets show that our EA$^2$E model brings improvement against previous methods.

%% file: figures/example.tex
\begin{figure}[!ht]
    \centering
    \includegraphics[width=0.99\linewidth]{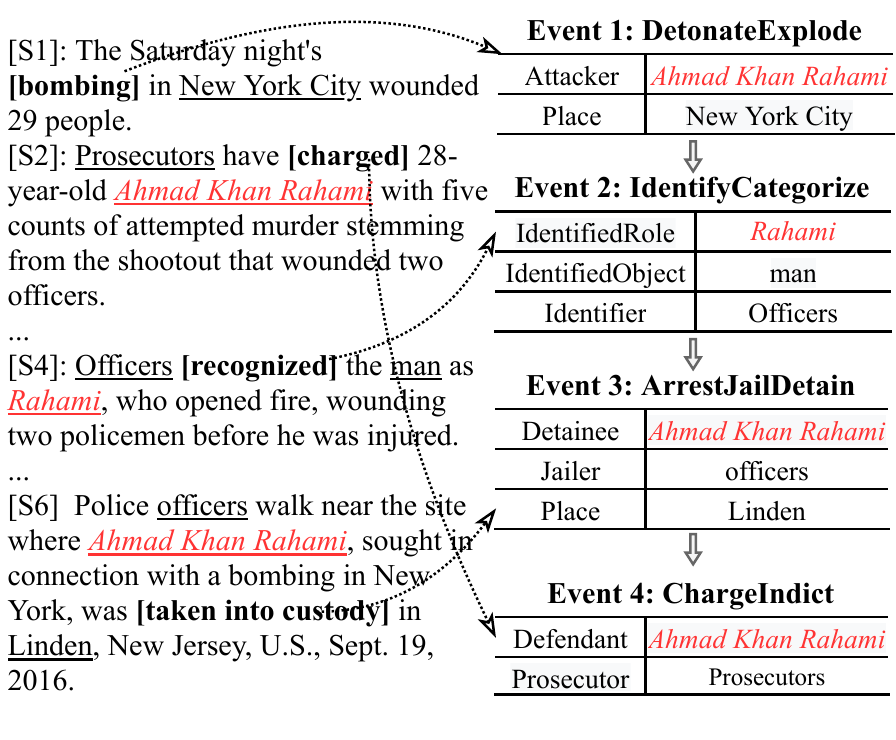}
    \caption{
    Examples of extracting arguments for multiple events in one document. 
    The casual relation between the \textit{Arrest} event and the \textit{Detonate} event puts their arguments under consistency constraints: \textit{Ahmad Khan Rahami}, the detainee in Event 3, is very likely to be the attacker in Event 1.
    Sentence-level models tend to miss the cross-sentence attacker argument in Event 1.
    }
    \label{fig:example}
\end{figure}

%% file: 300method.tex
\section{Event-Aware Argument Extraction}

\input{figures/framework}

Motivated by the observation that introducing event-event relations benefits the consistency of event argument extraction, 
we propose to incorporate implicit event-event relations with an Event-Aware Argument Extraction (EA$^2$E) model.
As shown in Figure~\ref{fig:model}, EA$^2$E contains alignment-enhanced training and iterative inference with self-augmented context.
When extracting the arguments for a target event, the context is augmented by labeling the arguments from neighboring events.
During training, an auxiliary training loss pulls close the event argument representations under the regular context and self-augmented context.
During inference, iterative inference encourages event awareness by using the extraction arguments from the last inference iteration as inputs.


\subsection{Base Encoder-Decoder Model}

Following~\cite{DBLP:conf/naacl/LiJH21}, we formulate event argument extraction as a conditional generation task
under the assumption that there exists a pre-defined event ontology describing each event type with an unfilled template with argument placeholders.
For example, the template for \textit{Attack} events is \textit{<arg> detonated or exploded  <arg>  explosive device using  <arg>  to attack  <arg>  target at  <arg>  place}.
Formally, given a document context $c$ and an event trigger $x$ with template $t$, the task is to extract a set of arguments 
$y = \{a_1, a_2,...a_n\}$, where each $a_i$ corresponds to a role predefined in the ontology.

We base our model on BART~\cite{DBLP:conf/acl/LewisLGGMLSZ20}, an encoder-decoder pretrained model.
The input sequence is the concatenation of the document context and an event template, constructed as \textit{<s> template </s> </s> context </s>}.
The output is a filled-in template, where the tokens are all selected from the input context or template.

The model parameter $\theta$ is trained by minimizing the argument extraction loss, the conditional probability over all instances:
\begin{equation}
    \mathcal{L}_E = - \sum log p_\theta (y|x,t,c)
\end{equation}

\subsection{Self-augmented Context}

We refer event awareness as the implication of event-event relations and reach this goal by labeling the arguments of other neighboring events in the context.
Given the arguments of the neighboring events $\{j\in \mathcal{N}_i\}$, which have small token-wise distances to the target event $i$, we augment the regular context $c$ by labeling them with \textit{<tag>}:
\begin{equation}
    c^{\prime}_i = \mu(c_i, \{y_{j}\}),
\end{equation}
where $\mu$ is the tagging operation,  and $y_j$ is the arguments of event $j$.

For example in Figure~\ref{fig:example}, when extracting arguments for the target event \textit{bombing} (Event 1), the augmented context is ``The Saturday night's <trg> bombing <trg> in New York City, wounded 29 people. \textit{<tag> Prosecutor <tag>} Prosecutors have charged 28-year-old \textit{<tag> Defendant <tag>} Ahmad Khan Rahami...", where the two tags highlight the arguments of Event 4.

\subsection{Alignment-enhanced Training}

An encoder is considered consistent when it is able to understand and encode the underlying relation between events into the text representations.
Therefore, we propose to enhance the encoder with an auxiliary training loss $\mathcal{L}_T$ that pulls close the argument representation distributions under regular context $c$ and under augmented context $c^{\prime}$. 
During training, $c^{\prime}$ is constructed by tagging the ground-truth arguments of neighboring events.

\begin{equation}
    \mathcal{L}_T = \sum  \|p(a|c),p(a|c^{\prime})\|_2
\end{equation}

The final training loss is a weighted sum of argument extraction losses ( $\mathcal{L}_E$ for regular context $c$ and $\mathcal{L}_{E^\prime}$ for augmented context $c^\prime$) and alignment-enhanced loss ($\mathcal{L}_T$) with weights $\alpha$ and $\beta$ :
\begin{equation}
    \nonumber 
    \mathcal{L} =  \mathcal{L}_E + \alpha \mathcal{L}_{E^\prime} + \beta \mathcal{L}_T
\end{equation}

\subsection{Iterative Inference}

Iterative inference explicitly introduces event awareness by utilizing extracted results in multiple inference iterations.
In the first iteration, for each target event trigger $i$, the model obtains the predicted results $y^1_i$ given the regular context $c_i^1$.
For the $k$-th iteration of inference, for each event trigger $i$ the context $c^{k}_i$ is augmented by labeling the extracted arguments $\{y^{k-1}_j\}$ of neighboring events $\{j\in \mathcal{N}_i\}$ from the $(k-1)$-th iteration.

\begin{equation}
    \nonumber
    c^{k}_i = \mu (c, \{y^{k-1}_j\})
\end{equation}

%% file: figures/framework.tex
\begin{figure*}[ht]
    \centering
    \includegraphics[width=1.0\linewidth]{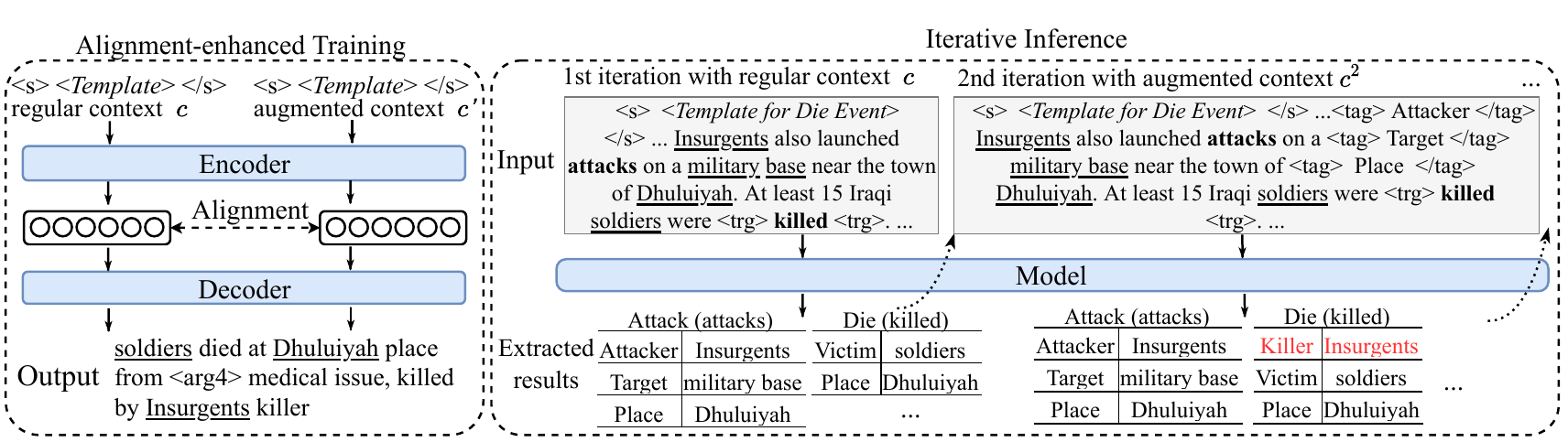}
    \caption{
    Our proposed Event-aware Argument Extraction model
    with alignment-enhanced training and iterative inference.
    During training, an auxiliary training loss aligns the event argument representations under regular context and augmented context.
    During inference, the context is augmented with results from the last iteration.
    }
    \label{fig:model}
\end{figure*}


%% file: 400experiment.tex
\section{Experiments}

\input{tables/wikievents}

\input{tables/result_ace}

\subsection{Datasets and Baselines}

We evaluate our proposed method on WIKIEVENTS~\cite{DBLP:conf/naacl/LiJH21} dataset and ACE 2005 dataset\footnote{\url{https://www.ldc.upenn.edu/collaborations/past-projects/ace}}.
Following previous work~\cite{DBLP:conf/acl/LiJH13}, we consider an argument span to be correctly identified when its offsets match any of the reference arguments of the current event (i.e., \textbf{Argument Identification}), and to be correctly classified when its role matches (i.e., \textbf{Argument Classification}).
We report the argument extraction performance in terms of Head Word F1 and Coreferential Mention F1. 
For the latter, full credit will be given when the extracted argument is coreferential with the gold-standard argument.

We compare EA$^2$E with document-level BART-Gen~\cite{DBLP:conf/naacl/LiJH21}, sentence-level ONEIE~\cite{DBLP:conf/acl/LinJHW20} and BERT-CRF~\cite{DBLP:journals/corr/abs-1904-05255}.

\subsection{Implementation Details}

We implement our models with Huggingface~\cite{DBLP:journals/corr/abs-1910-03771}.
We train each model for 4 epochs with a batch size of 4 for baselines and 2 for EA$^2$E.
The model is optimized with the Adam optimizer with a learning rate of 3e-5, $\alpha$ = 1 and $\beta$=0.5. 
We define event neighborhood as trigger distance less than 40 tokens.
For inference, the maximum number of iterations is 3.
For numerical consistency, all experiment results are averaged across three random runs.
The hyper-parameters are selected by grid search based on model performance on development set. \footnote{$\beta$ is chosen from $\{0.1, 0.5, 1\}$, the trigger distance is chosen from $\{20,40,60,80,100\}$, and the learning rate is chosen in $\{3e-5, 5e-5\}$.}
On average it takes approximately one hours to train a model until converge with one Tesla P100 GPU with 16GB DRAM. 

\subsection{Results and Analysis}

\input{tables/improved_cases}

Table~\ref{tab:wikievent} and Table~\ref{tab:ace} show that our proposed EA$^2$E consistently performs better than strong baseline methods across datasets and evaluation metrics. In general document-level methods have better performance, especially in terms of recall, because sentence-level methods are more likely to miss cross-sentence arguments.

Alignment-enhanced training brings a significant improvement over BART-Gen but comes with higher training costs since the inputs are doubled.

Iterative inference brings unstable improvement.
Figure~\ref{fig:iter} shows that more iterations brings higher performance only to a certain range.
Since the only differences among iterations are their inputs, we conclude that labeling the arguments of other events helps the model extract the arguments of the current event. 
The upper bound of this improvement is limited by the error propagation in the augmented context.

\input{figures/iter}

\textbf{Qualitative Analysis} 
Table~\ref{tab:improved_cases} presents some representative examples. 
BART-Gen incorrectly assigns \textit{Tsarnaev} to the Target role, and \textit{police} to the attacker role in the first example.
It also misses the killer \textit{brothers} in the second example and the attacker \textit{Laden} in the third example.
The second example shows the advantage of the Alignment-enhanced Training component in EA$^2$E, which helps extract the killer argument. 
The third example shows how Iterative Inference works with the augmented input: 
The tagged attacker in the neighboring \textit{bombing} event is also the attacker in the target \textit{attack} event.

\textbf{Remaining Challenges} 
Though effective, iterative inference may propagate errors among iterations.
In addition, the success of event awareness relies on the assumption that events in a neighborhood defined by trigger distance are inter-related to the target event.
However, this assumption is not always held true in the case that distant events bring redundant information.
It is not necessarily hurting the information but it brings noise by incorrectly implying the relations between the distant events and the target event.

%% file: tables/wikievents.tex
\begin{table*}[htbp]
\small
\centering
\setlength\tabcolsep{4.0pt}

\begin{tabular}{lcccccccccccc}

\toprule[1pt]
\multirow{3}{*}{\textbf{Model}}	&	\multicolumn{6}{c}{\textbf{Argument Identification}}   & \multicolumn{6}{c}{\textbf{Argument Classification}}\\

&\multicolumn{3}{c}{\textbf{Head Match}}&\multicolumn{3}{c}{\textbf{Coref Match}}&\multicolumn{3}{c}{\textbf{Head Match}}&\multicolumn{3}{c}{\textbf{Coref Match}} \\

&P&R&F1&P&R&F1&P&R&F1&P&R&F1\\
\hline
BERT-CRF &72.66&53.82&61.84&74.58&55.24&63.47&61.87&45.83&52.65&63.79&47.25&54.29\\
ONEIE & 68.16&56.66&61.88&70.09&58.26&63.63&63.46&52.75&57.61&65.17&54.17&59.17\\
BART-Gen

&70.43&71.94&71.18&71.83&73.36&72.58&65.39&66.79&66.08&66.78&\textbf{68.21}&67.49\\
\hline

\textbf{EA$^2$E}&
76.51&\textbf{72.82}&\textbf{74.62}&77.69&\textbf{73.95}&\textbf{75.77}&70.35&\textbf{66.96}&\textbf{68.61}&71.47&68.03&\textbf{69.70}\\
EA$^2$E w/o AT&
\textbf{77.26}&71.23&74.12&\textbf{78.61}&72.47&75.42&\textbf{71.10}&65.54&68.21&\textbf{72.25}&66.61&69.32\\
EA$^2$E w/o II & 
75.96&72.29&74.07&77.13&73.42&75.22&69.61&66.25&67.89&70.72&67.32&68.97\\
\bottomrule[1pt]
\end{tabular}

\caption{
Performance (\%) on WIKIEVENTS dataset. \textbf{AT}: Alignment-enhanced Training. \textbf{II}:Iterative Inference.
}
\label{tab:wikievent}
\end{table*}%

%% file: tables/result_ace.tex
\begin{table*}[htbp]
\small
\centering

\setlength\tabcolsep{4.0pt}

\begin{tabular}{lcccccccccccc}

\toprule[1pt]
\multirow{3}{*}{\textbf{Model}}	&	\multicolumn{6}{c}{\textbf{Argument Identification}}   & \multicolumn{6}{c}{\textbf{Argument Classification}}\\

&\multicolumn{3}{c}{\textbf{Head Match}}&\multicolumn{3}{c}{\textbf{Coref Match}}&\multicolumn{3}{c}{\textbf{Head Match}}&\multicolumn{3}{c}{\textbf{Coref Match}} \\

&P&R&F1&P&R&F1&P&R&F1&P&R&F1\\
\hline
BERT-CRF&65.77&51.04&57.48&67.11&52.08&58.65&56.82&44.10&49.66&57.72&44.79&50.44 \\
ONEIE&
63.33&61.46&62.38&65.12&63.19&64.14&58.50&56.77&57.62&60.11&58.33&59.21\\
BART-Gen
&70.00&73.84&71.87
&71.37&75.29&73.27
&65.72&69.33&67.47
&66.76&70.43&68.54\\
\hline

\textbf{EA$^2$E}
&\textbf{74.54}&74.88&\textbf{74.71}
&\textbf{75.81}&76.16&\textbf{75.98}
&\textbf{71.83}&\textbf{72.16}&\textbf{72.00}
&\textbf{72.98}&\textbf{73.32}&\textbf{73.15}\\
EA$^2$E w/o AT 
&73.95&74.25&74.10
&75.28&75.58&75.43
&70.78&71.06&70.92
&72.05&72.34&72.19\\
EA$^2$E w/o II 
&74.36&\textbf{75.00}&74.68
&75.56&\textbf{76.22}&75.88
&71.49&72.11&71.80
&72.58&73.20&72.89\\
\bottomrule[1pt]
\end{tabular}

\caption{Performance (\%)  on ACE2005 dataset.}
\label{tab:ace}
\end{table*}%

%% file: tables/improved_cases.tex
\begin{table*}[htbp]
\small
\centering
\setlength\tabcolsep{4.0pt}

\resizebox{\textwidth}{!}{

\begin{tabular}{lp{3cm}p{3cm}p{3cm}p{3cm}}

\toprule[1pt]
 & \textbf{Gold} & \textbf{BART-Gen} & \textbf{EA$^2$E w/o II} & \textbf{EA$^2$E}\\
\midrule


\textbf{Input} & \multicolumn{4}{l}{Dzhokhar \uwave{Tsarnaev} visits Silva and borrows the Ruger pistol — the \uline{gun} that was later used to kill MIT police officer } \\
& \multicolumn{4}{l}{Sean Collier and during the  \textbf{shootout}  with \uline{police} in \uwave{Watertown}.} \\

\textbf{Output} & Target: police & \textcolor{red}{Target: Tsarnaev} & Target: police & Target: police\\
& Instrument: gun & \textcolor{red}{ Attacker: police} & Attacker: Tsarnaev & Attacker: Tsarnaev \\
& & Place: Watertown & Place: Watertown & Place: Watertown \\
\midrule


\textbf{Input} & \multicolumn{4}{l}{The \uline{brothers} allegedly set off two bombs alongside the    Boston Marathon    course,  \textbf{killing}  three \uline{people} and }  \\
& \multicolumn{4}{l}{injuring  264.} \\
\textbf{Output} & Killer: brothers & Victim: people & Killer: brothers & Killer: brothers \\
& Victim: people &   & Victim: people & Victim: people\\
\midrule


\textbf{Input} & \multicolumn{4}{l}{Osama bin \uline{Laden} is charged with masterminding the 1998 bombings of two U. S. embassies in  East Africa, believed }  \\
& \multicolumn{4}{l}{to have had a role in the October 2000  \textbf{attack}  on the USS \uline{Cole} in the \uwave{Yemeni} \uline{port} of Aden. } \\

\textbf{Augmented} &  \multicolumn{4}{l}{<tag>  Attacker  </tag>  Osama bin \uline{Laden} is charged with masterminding the 1998 bombings of two U. S.  <tag>  } \\
\textbf{Input}& \multicolumn{4}{l}{Target  </tag>  embassies in  <tag>  Place  </tag>  East Africa, believed to have had a role in the October 2000  } \\
& \multicolumn{4}{l}{\textbf{attack} on the USS \uline{Cole} in the \uwave{Yemeni} \uline{port} of Aden. }\\

\textbf{Output} & Target: Cole  & Target: Cole & Target: Cole & Target: Cole  \\
& Target: port & Place: Yemeni& Place: Yemeni &Place: Yemeni \\
& Attacker: Laden &&& Attacker: Laden \\

\bottomrule[1pt]
\end{tabular}
}

\caption{Examples of extracted arguments from BART-Gen, EA$^2$E w/o II, and EA$^2$E. 
We label \textbf{target event mention} with bold, \uline{gold arguments} with underlines, \uwave{correct but not annotated arguments} with waves, and \textcolor{red}{incorrect arguments} with red.
In the third example we also present the augmented input for Iterative Inference, in which the arguments of the bombing event are tagged.
}
\label{tab:improved_cases}
\end{table*}

%% file: figures/iter.tex
\begin{figure}[htb]
    \centering
    \includegraphics[width=0.90\linewidth]{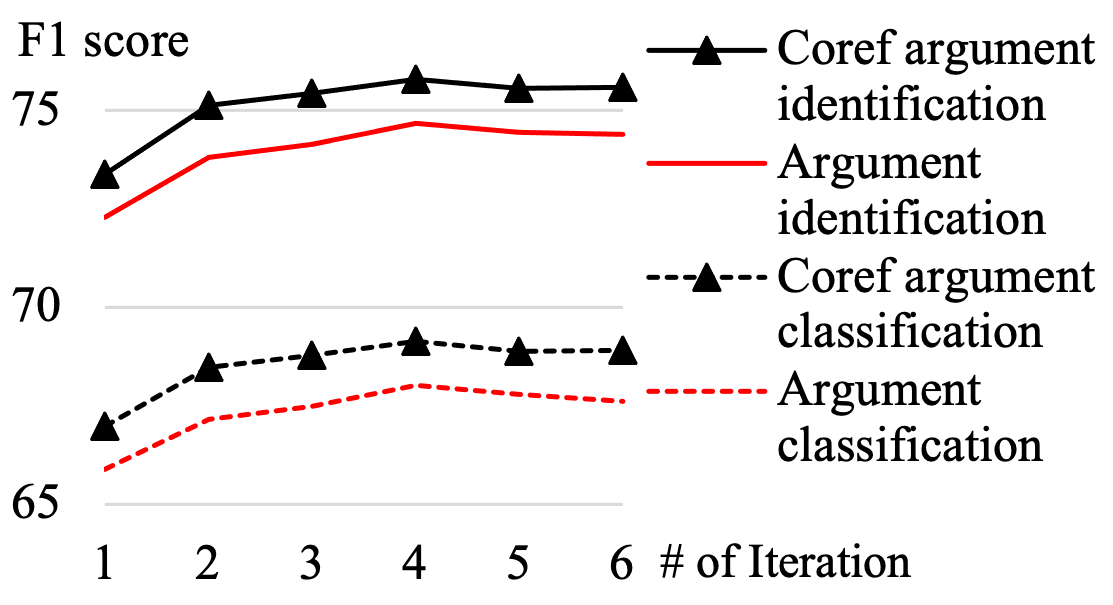}
    \caption{F1 score (\%) of EA$^2$E for different iterations in a single run on WIKIEVENTS. 
    }
    \label{fig:iter}
\end{figure}

%% file: 500related.tex
\section{Related Work}

Sentence-level argument extraction approaches, where the event trigger and its arguments are usually located within a single sentence, cannot handle the cross-sentence trigger-argument distribution and the existence of multiple events within one document.
Though recent attempts on document-level argument extraction have gone beyond sentence boundaries, they either focus  on one-event-per-document setting, or model each event independently.
The most related work to ours is~\cite{DBLP:conf/naacl/LiJH21}, which formulates the task as conditional generation following event templates and extracts arguments for each event independently, while our work focuses on the consistency among arguments for different events.
The methods in~\cite{DBLP:conf/eacl/DuRC21} and~\cite{DBLP:conf/acl/DuC20} are designed for Role-filler Entity Extraction (REE) task under the assumption that one generic template is produced for each document, while our work focuses on extracting arguments for multiple events for each document.
\cite{DBLP:conf/acl/YangS000W20} introduces Parallel Prediction Network that generates all possible events in parallel based on the document-aware representations, while we adopt a generative framework.
\cite{DBLP:conf/acl/XuLLC20} model the whole document as graphs and capture the interdependency among events by tracking the extracted events with a global memory, while we introduce event awareness for interdependency without external memory modules.

%% file: 600conclusion.tex
\section{Conclusion}
We introduce Event-Aware Argument Extraction (EA$^2$E) model to improve self-contained consistency in document-level event argument extraction.
We conclude that in iterative inference brings higher performance only to a certain range of iterations and alignment-enhanced training brings significant improvement with costs.